\documentclass[10pt,twocolumn,letterpaper]{article}

\usepackage{wacv}              

\usepackage{graphicx}
\usepackage{amsmath}
\usepackage{amssymb}
\usepackage{booktabs}
\usepackage[accsupp]{axessibility} 

\usepackage[pagebackref,breaklinks,colorlinks]{hyperref}

\usepackage[capitalize]{cleveref}
\crefname{section}{Sec.}{Secs.}
\Crefname{section}{Section}{Sections}
\Crefname{table}{Table}{Tables}
\crefname{table}{Tab.}{Tabs.}

\def\wacvPaperID{603} 
\def\confName{WACV}
\def\confYear{2025}

\usepackage[dvipsnames,table]{xcolor}

\usepackage{float}
\usepackage{enumitem}

\usepackage{orcidlink}
\usepackage{makecell}

\usepackage{tabularx}
\usepackage{adjustbox}
\usepackage{graphics}
\definecolor{DarkCyan}{rgb}{0.4, 0.85, 0.8}

\newcommand{\boldparagraph}[1]{\vspace{0.1em}\noindent{\bf #1} }

\newcommand{\added}[1]{#1}

\begin{document}

\title{XR-MBT: Multi-modal Full Body Tracking for XR through Self-Supervision with Learned Depth Point Cloud Registration}

\author{Denys Rozumnyi \and Nadine Bertsch \and Othman Sbai \and Filippo Arcadu \and Yuhua Chen \and Artsiom Sanakoyeu  \qquad Manoj Kumar \qquad   Catherine Herold \qquad  Robin Kips \\  \\
Meta Reality Labs Zurich
}

\maketitle

\begin{abstract}
  Tracking the full body motions of users in XR (AR/VR) devices is a fundamental challenge to bring a sense of authentic social presence. 
Due to the absence of dedicated leg sensors, currently available body tracking methods adopt a synthesis approach to generate plausible motions given a 3-point signal from the head and controller tracking. 
In order to enable mixed reality features, modern XR devices are capable of estimating depth information of the headset surroundings using available sensors combined with dedicated machine learning models.
Such egocentric depth sensing cannot drive the body directly, as it is not registered and is incomplete due to limited field-of-view and body self-occlusions. 
For the first time, we propose to leverage the available depth sensing signal combined with self-supervision to learn a multi-modal pose estimation model capable of tracking full body motions in real time on XR devices.
We demonstrate how current 3-point motion synthesis models can be extended to point cloud modalities using a semantic point cloud encoder network combined with a residual network for multi-modal pose estimation.
These modules are trained jointly in a self-supervised way, leveraging a combination of real unregistered point clouds and simulated data obtained from motion capture.
We compare our approach against several state-of-the-art systems for XR body tracking and show that our method accurately tracks a diverse range of body motions.
XR-MBT tracks legs in XR for the first time, whereas traditional synthesis approaches based on partial body tracking are blind.

\end{abstract}

\begin{figure}
    \centering 
    \includegraphics[width=0.48\textwidth]{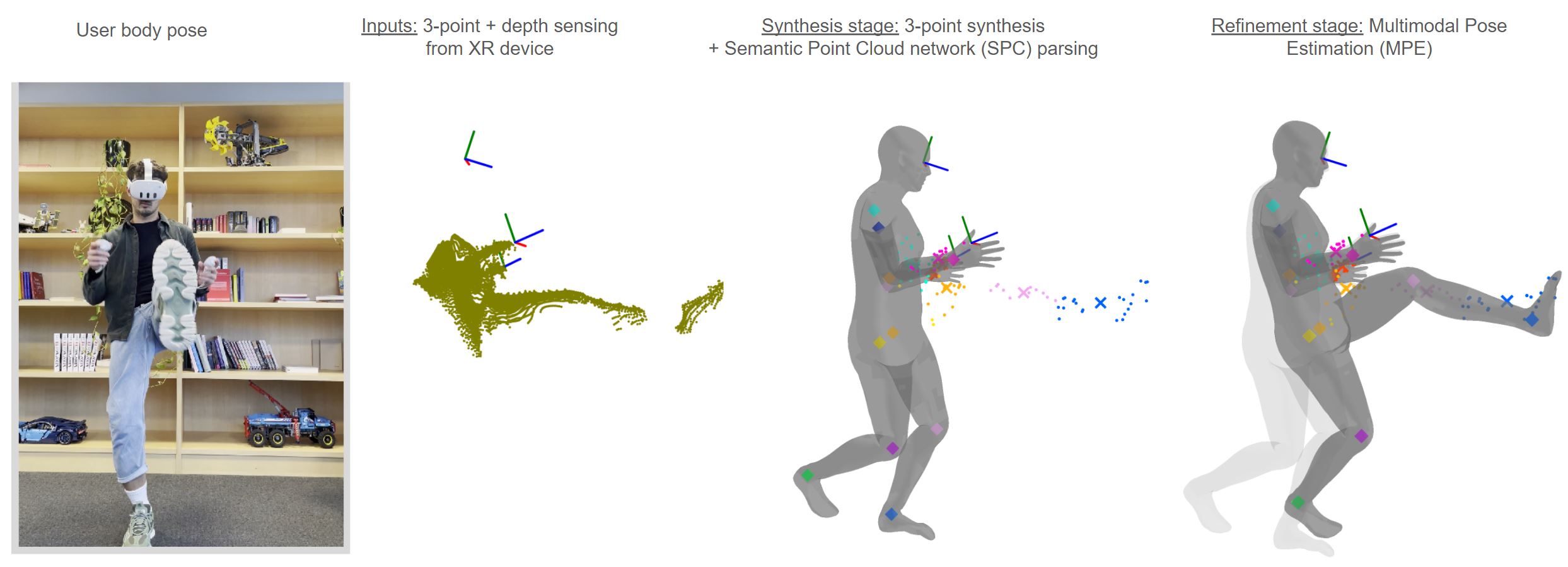}
    \vspace{-2.1em}
    \captionof{figure}{XR-MBT uses 3-points (head/wrists) and depth sensing from the XR device to learn real-time multi-modal body tracking. First, the synthesis stage generates a plausible body pose using the 3-Point signal. Then, the semantic point cloud network registers the body point cloud, and our self-supervised residual multi-modal pose estimation network predicts the refined body pose.}
    \label{fig:header}
\end{figure}

\begin{figure*}
\centering
\includegraphics[width=0.9\linewidth]{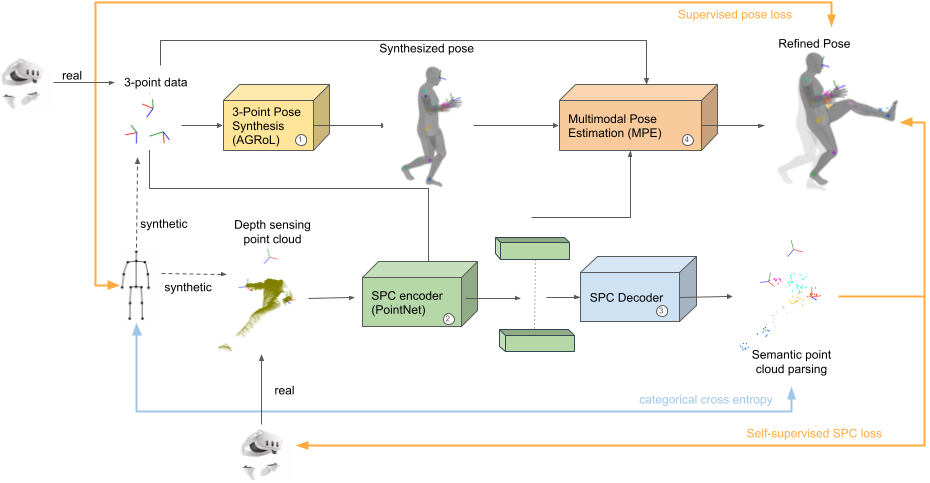}
\vspace{-1em}
\caption{\label{fig:training_archi} \textbf{XR-MBT architecture.}
First, we use AGRoL~\cite{du2023agrol} to synthesize the initial pose (yellow).
Second, we process the point cloud by the Semantic Point Cloud (SPC) encoder (green) to generate point features. 
Third, the SPC decoder (blue) generates the probability of mapping of every point to a body joint, which is used for self-supervised learning.
Last, the Multi-modal Pose Estimation (MPE) network (orange) estimates the final body pose by combining all modalities.
MPE and SPC networks are jointly trained with a combination of Mocap and unlabeled depth data.}
\end{figure*}

\section{Introduction}
\label{sec:intro}

The creation of immersive XR experiences requires bringing realistic body representations for users to AR/VR devices. The full body motions of the users must be accurately represented to create authentic social experiences or interactions between users and their environment. This is also required to create a smooth user experience by accurately reproducing the whole body behavior in AR/VR applications, showing the virtual body as close as possible to the real one. 
Current XR body tracking methods focus on the "3-point" task, where signals from the head and controller tracking systems are leveraged to predict full body poses \cite{winkler2022questsim,du2023agrol,shi2023phasemp,zheng2023realistic}.
Due to the absence of lower-body sensors in standalone XR devices,
these approaches heavily rely on synthesis and generative models to reconstruct a plausible full-body motion. 
While recent methods explore egocentric vision-based body tracking~\cite{rhodin2023egocap, jiang2017seeinginvisibleposes, yuan20183degoviaimitation}, they require datasets collection with high-quality ground-truth, making them prohibitively expensive to execute at a large scale. 
This calls for the creation of novel methods to relax the need for labeled data for lower body tracking. 

Due to the rapid development of augmented reality systems, depth sensing has become a core component of modern XR devices. In fact, real-time depth estimation is at the center of multiple essential systems for XR devices, from Passthrough rendering \cite{chaurasia2020passthrough, xiao2022neuralpassthrough} to sensing user environments for safety and solving occlusion when rendering virtual objects in the user environment. Figs.~\ref{fig:header} and~\ref{fig:depth_data_demo} show such egocentric point cloud data obtained from an XR device. 

However, these \textit{egocentric} point clouds present characteristics that prevent them from being used directly in body tracking, which we address in this work. 
First, the point cloud is \textit{unregistered} as correspondences with body parts cannot be immediately identified to drive the body pose estimation. 
Second, the egocentric body point cloud is \textit{incomplete}. 
As the depth is captured by headset sensors located on the user's head, most body parts are only partially visible. 
Visibility is limited through the dependency of the sensor's field-of-view (FOV) on the user's head pose and also through self-occlusions. 
This signifies that body tracking still needs to rely on synthesis for body parts not captured by the sensing signal. 
Finally, egocentric point clouds have a \textit{characteristic noise} due to the sensor placement specificity and the depth-sensing method, which require the use of real data to train body tracking models to learn sensor domain adaptation and generalize to diverse body poses.

For the first time in the XR body tracking field, we propose leveraging depth sensing of XR devices to introduce multi-modal full-body tracking. 
The \textbf{contributions} can be summarized as follows:
\begin{itemize}[itemsep=0.1pt,topsep=3pt,leftmargin=*]
  \item We introduce XR-MBT (Fig.~\ref{fig:training_archi}), a new multi-modal architecture to extend 3-point-based motion synthesis models to leverage depth sensing signal available in XR devices, allowing full body tracking in real-time.
  \item We propose a novel self-supervised method to train a multi-modal motion estimation network from unregistered point cloud data using a Semantics Point Cloud (SPC) loss leveraging a trained SPC network. 
  This allows us to learn the robustness of the depth sensing system from real data while relaxing the need to collect expensive and complex ground truth data.
  \item  We collect an egocentric body depth dataset with head and wrist tracking information, which can be used to explore a new research area around multi-modal body tracking for XR and increase the diversity of tracked motions. 
  \item We demonstrate that our method can be applied to VR devices, achieving real-time tracking of the full body for XR for the first time, and we demonstrate the different elements of our method in an ablation study under a controlled synthetic environment, as well as using real data. 
\end{itemize}

\section{Related work}
\label{sec:related}
\subsection{Motion synthesis for XR}
Human motion synthesis is a key area in computer vision, with applications spanning animation, AR/VR and digital humans. The goal is to synthesize motions that are not only realistic but also expressive and diverse, catering to various tasks such as motion interpolation/prediction~\cite{martinez2017human}, denoising~\cite{kim2019human}, and text-to-motion generation~\cite{zhang2024motiondiffuse, guo2022generating}. 
Given the temporal nature of human motion, models like RNNs have been successfully employed for this task~\cite{holden2016deep, kim2019human, fragkiadaki2015recurrent}.
In recent years, generative models like VAEs~\cite{aliakbarian2020stochastic, martinez2017human} and GANs~\cite{hernandez2019human} have been extensively explored for their ability to generate high-quality, plausible motion data. The latest advancements have seen the emergence of diffusion models in motion synthesis~\cite{yuan2023physdiff, zhang2024motiondiffuse, tevet2022human}, offering a new paradigm for generating motion by gradually transforming random noise into structured motion data. 

Motion synthesis for XR has long been recognized as a key component of immersive XR experiences. 
The XR user wears a head-mounted device and hand controllers equipped with inertial measurement units (IMUs) that are used together with a vision system to track the position and orientation of the head and hands. 
This signal is typically referred to as the "3-Point" input. Predicting the full body pose from a 3-Point sequence represents an under-constrained problem with sparse inputs. 
\cite{huang2018deepinertialposer, yang2021lobstr, jiang2022inertialposer} propose to address the motion synthesis problem with a bidirectional RNN 
by adding IMUs placed on the pelvis and lower body on top of the 3-Point input. 
\cite{dittadi2021smplposesfromhmd, aliakbarian2022flag, jiang2022avatarposer} tackled the problem of egocentric pose estimation from just 3-Point input using VAEs.
In \cite{winkler2022questsim}, a deep reinforcement learning model with physics simulation was proved capable of synthesizing full body poses from either 3-Point input or the device pose alone. 
A method to generate a full body pose from 3-Point was also proposed using physics simulations in \cite{ye2023neural3points}. 
Some egocentric pose datasets with depth maps have been proposed such as xR-EgoPose~\cite{tome2019xr, tome2020self} and UnrealEgo~\cite{akada2022unrealego}.
A different direction to increase the number of tracked joints consists of estimating full-body pose from egocentric cameras mounted on the head~\cite{rhodin2023egocap, jiang2017seeinginvisibleposes, yuan20183degoviaimitation, cha2018towardsmobile3dbody, ng2020you2me, zhao2021egoglass, akada2022unrealego, jiang2017seeinginvisibleposes, wang2023fisheyevitanddiffusion, kang2023ego3dpose}.
However, none of these studies focused on the specificity of the XR use case, where the availability of the 3-Point signal, in addition to the headset camera and depth sensors, requires multi-modal body tracking models. 

Building multi-modal body tracking models requires large-scale ground-truth data collection to train models that can generalize to diverse body poses and environments for each sensor set, which leads to prohibitive costs.
Unfortunately, such datasets dedicated to XR body tracking do not yet exist. To the best of our knowledge, no existing study addresses the problem of estimating full-body pose by fusing multi-modal inputs provided by an XR device.

\subsection{Depth sensing for XR}

Depth on AR/VR devices can be obtained based purely on vision, \eg from stereo cameras \cite{chaurasia2020passthrough}. 
Nevertheless, recent advances in machine-learning-based depth estimation, \eg CRE-Stereo \cite{li2022practical}, made ML-based depth on the device a standard approach~\cite{Meta_blog}. 
Alternative but complementary approaches are physical depth sensors such as dToF, which are also available on most smartphones today. 
For most of the applications using depth on XR devices, a strong constraint is to provide high-frequency and low-latency depth, which matches body-tracking needs. 
Thus, the choice of the depth sensing approach is determined given an aggregation of constraints to balance (such as power, hardware, frequency, quality, resolution, \etc) and can vary between different devices. 
In this work, we propose a method that is independent of the depth sensing technology.
We only postulate that the XR device contains depth sensing technology that returns an unstructured point cloud. 
We also assume that the depth sensor is attached to the head-mounted device with a limited field of view of the human body.

\subsection{3D Human pose estimation from depth}

The rapid evolution of deep learning techniques for processing 3D point cloud data has significantly enhanced the capabilities in various fields such as robotics, autonomous vehicles, and computer vision applications. 
Recent studies have delved into key tasks like 3D object classification, tracking, pose estimation, segmentation, and point cloud completion, showcasing the strides made in the development and application of these methods. \cite{ding2023recent, xu2021review} provide a comprehensive review of these advancements, highlighting the historical progression, model architectures, learning algorithms, and the challenges that still persist. 
Focusing on real-time performance,~\cite{biswas2019fast, schnurer2019real} propose a multi-stage architecture by first estimating a 2D pose from RGB or depth maps followed by a 3D pose estimation.

\added{Previous point cloud-based human pose estimation methods have proposed to combine point clouds with IMU for pose estimation~\cite{r4_paper1, r4_paper2, r4_paper3}.}
Furthermore, \cite{vasileiadis2019multi} proposes a multi-person 3D pose estimation system with 3D CNNs to overcome the limitation of using depth maps as single-channel information. 
The advancement of sequential 3D human pose and shape estimation from point clouds highlights the integration of spatial-temporal features to recover dynamic 3D body models accurately. 
\cite{d2021refinet} introduces novel multi-stage architecture RefiNet to progressively use 2D keypoints, depth maps and point clouds refining 3D human pose. 
These approaches tackle the inherent challenges of 3D pose estimation, including the ambiguity and complexity of human poses, the difficulty of annotating 3D joints in point clouds, and the demand for multi-person scenes. 
The introduction of hierarchical regression frameworks for detailed parametric model estimation~\cite{wang2020sequential, wang2021parametric} from partial point clouds addresses the challenges related to the variability in human shapes and clothing details. 
Although not from egocentric views, \cite{zhang2020weakly} proposes a weakly supervised adversarial learning approach to generate 3D human pose leveraging both depth maps and point clouds. 
To the best of our knowledge, predicting full-body pose from egocentric depth and point clouds remains an understudied problem.

\section{Method}
\label{sec:method}

\begin{figure}[t]
\centering
\includegraphics[width=0.99\linewidth]{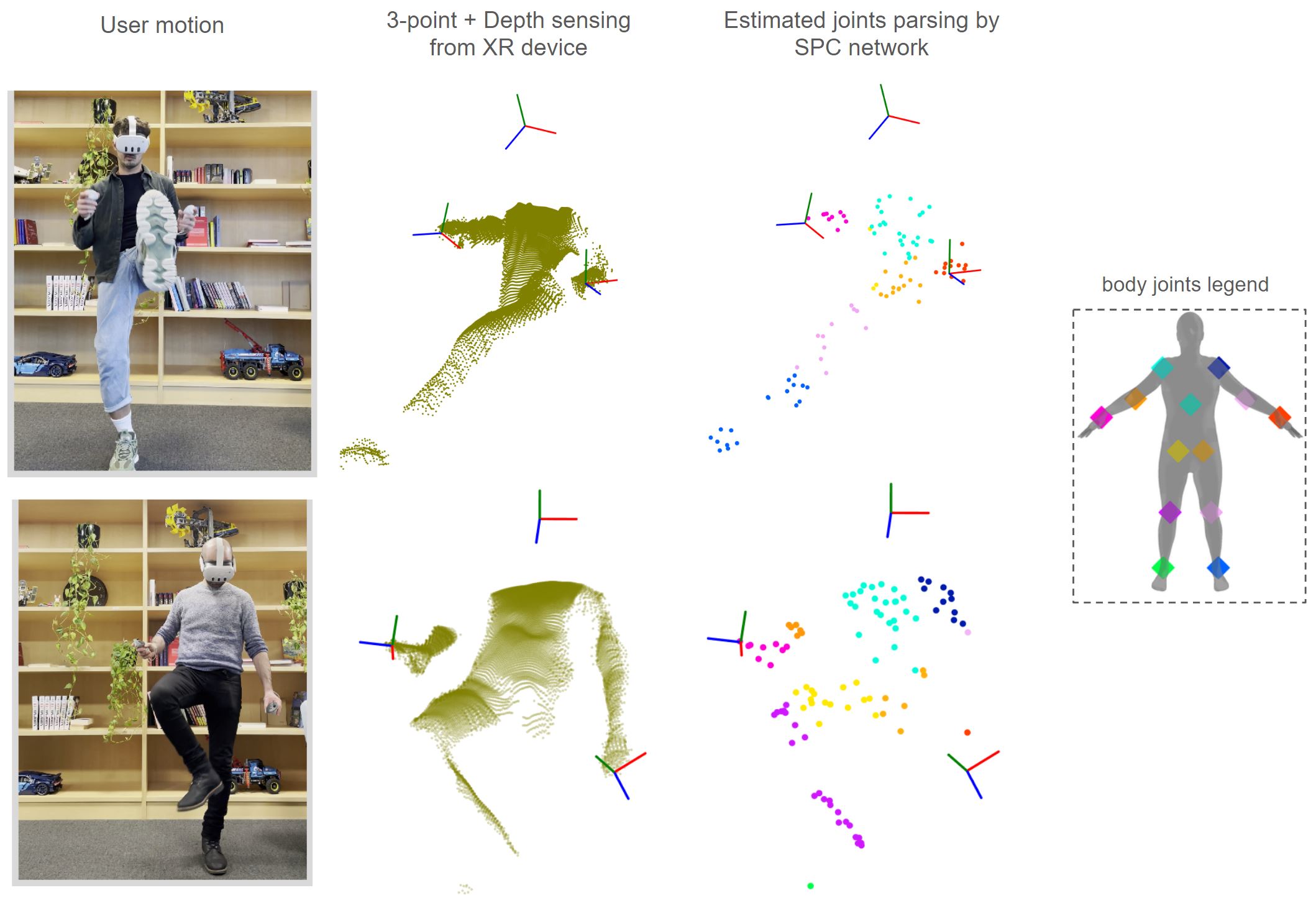}
\vspace{-1em}
\caption{\label{fig:depth_data_demo} An example of real 3-Point and point cloud data accessible from XR devices and the corresponding output of our semantic point cloud (SPC) network. 
Even though the body point cloud is partial due to self-occlusions and outside the field of view, our SPC network learns to correctly identify the various body point joints by leveraging the 3-Point information. This indicates that the SPC network learns a meaningful registration of the body point cloud to supervise the MPE network training.  }
\end{figure}

We propose for the first time to formulate the XR body tracking problem as a multi-modal task where the inputs are both the conventional 3-Point data, \eg from the headset and controllers,
and an unstructured point cloud obtained from the depth sensing system of the XR device. 
Such real inputs for diverse user motions are illustrated in Fig.~\ref{fig:depth_data_demo}.

In this section, we describe all components of XR-MBT, the proposed self-supervised multi-modal body tracking method. 
The Semantic Point Cloud (SPC) network learns to register an unstructured body point cloud to identify body joint information, while the Multi-modal Pose Estimation (MPE) residual network learns to refine body pose based on 3-Point data and SPC network embedding. 
First, we train both networks jointly on Mocap data with simulated synthetic point clouds.
Using a novel self-supervised loss with learned semantics, the multi-modal residual network is then fine-tuned on a dataset of unlabeled depth data to learn robustness to sensor noise.

\boldparagraph{Inputs.}
We assume the input is a sequence of length $N = 196$ of 3-point inputs $x^{1:N}$ and sampled point clouds $p^{1:N}$.
The 3-Point inputs consist of position, acceleration, rotation and rotational acceleration of each point.
We represent rotations as 6D rotations as introduced in~\cite{rot6d}. 
Thus, $x^{t} \in \mathbb{R}^{54}$.
Due to computational reasons and since point clouds at run-time have different numbers of points, we sample $P = 100$ points from each point cloud with probability proportional to their depth value to reinforce the presence of body parts such as feet that cover a smaller surface than the closer torso.
Hence, $p^{t} \in \mathbb{R}^{P \times 3}$.

\boldparagraph{Outputs.}
We denote the output of the synthesis network as 
\begin{equation}
    y^{1:N} = \text{synthesis}(x^{1:N}) \, ,
\end{equation}
with $y^{t} \in \mathbb{R}^{J \times 6}$, where for each of $J$ joints from a skinned body model representation (\eg SMPL~\cite{smpl}), we predict local joint rotations. 
The SPC network registers the point cloud by predicting joint class probabilities $l^{1:N}$, where $l^t \in \mathbb{R}^{P \times (J + 1)}$ represents the probability of each point in the point cloud to be assigned to one of $J$ joints or the background/outlier label, \eg for each point $i$, the following holds: $\sum_{j=1}^{J+1} l^t(i, j) = 1$.
Additionally, the SPC network predicts point cloud features $f^{1:N}$, where $f^{t} \in \mathbb{R}^{128}$, thus:
\begin{equation}
    (f^{1:N}, l^{1:N}) = \text{spc}(x^{1:N}, p^{1:N}) \, .
\end{equation}
Finally, the multi-modal pose estimation residual network predicts a final pose as an offset from local rotations $y^{1:N}$ of the synthesis network: 
\begin{equation}
    z^{1:N} = y^{1:N} + \text{mpe}(x^{1:N}, y^{1:N}, f^{1:N}) \, .
\end{equation}

\subsection{Semantic point cloud network}
\label{sec:spc}

As the egocentric point cloud obtained by the depth sensing system of the XR device is unlabeled, it cannot be directly leveraged to learn a multi-modal body tracking model.
The goal of the \textit{Semantic Point Cloud (SPC)} network is two-fold. 
First, it learns a compact encoding $f^{1:N}$ of point cloud features that are leveraged by a body tracking model. 
Second, it provides a semantic registration $l^{1:N}$ of the point cloud to identify the different body joints and label the point cloud. 
This registration is used to formulate a self-supervised loss described in Section~\ref{sec:loss}. 

The SPC network is composed of a PointNet~\cite{qi2017pointnet, qi2017pointnetplusplus} encoder-decoder architecture, and its training is illustrated in Fig.~\ref{fig:training_archi}. 
For each timestamp $t$ separately, it takes the sampled point cloud and the 3-Point data as input.
Indeed, the head and wrist poses provide an important motion trajectory context for interpreting the semantics of the point cloud. 
Furthermore, the point cloud positions are normalized with respect to the position of the depth sensor attached to the XR device. 
To represent the temporal context of the point cloud, the SPC decoder takes as input the local PointNet features from the SPC encoder and an average pooling of the global PointNet features over the previous timestamps of the sequence. 
The 3-Point data with rotations are also passed to the decoder through a skip connection to ensure that the headset and controller information is preserved. 
For each of the sampled points in the point cloud, the SPC decoder outputs the probability of mapping to a body joint.

The SPC network is trained on the Mocap data, where synthetic point clouds are generated by sampling points on the body mesh surface. 
The body shape is randomized to represent diverse users. 
To increase robustness in real-world scenarios, we add Gaussian noise with a standard deviation of 2 cm to all points. 
Then, we create outliers by adding Gaussian noise with a standard deviation of 20 cm to 2\% of all points in the point cloud.
The ground truth label for each point in the point cloud is defined as the closest joint in the Mocap ground truth body pose. 
If the closest joint is further away than a threshold of 10 cm, we label such points as background/outliers to represent sensor noise. 
The SPC encoder and decoder modules are jointly trained with all other modules using categorical cross-entropy loss between the predicted labeling $l^{1:N}$ and ground-truth labeling $\hat{l}^{1:N}$:
\begin{equation}
    \mathcal{L}_{ce} = \text{cross-entropy}(l^{1:N}, \hat{l}^{1:N}) \, .
\end{equation}

\subsection{Multi-modal pose estimation network}
\label{sec:mpe}
The proposed pose estimation method is composed of two stages: synthesis and refinement, and its training is described in Fig.~\ref{fig:training_archi}.
Because there is no guarantee that any body parts are in the field-of-view of the depth sensor, we initialize the body tracking system with a 3-Point pose synthesis to guarantee the plausibility of the pose. 
For this purpose, we reuse the AGRoL \cite{du2023agrol} generative diffusion-based model architecture for the synthesis stage. 
The role of the Multi-modal Pose Estimation residual network is to refine the synthesized pose using the point cloud information available from the depth sensing. 
The MPE network takes as inputs the 3-Point signal $x^{1:N}$, the point cloud information, represented as embedding $f^{1:N}$ of the SPC encoder, and the synthesis network output $y^{1:N}$.
Then, it predicts the final pose $z^{1:N}$ as an offset from the synthesis output.

The MPE network is trained on synthetic data, where point clouds are generated using a procedure similar to the SPC model training. 
The Mocap ground-truth pose $\hat{z}^{1:N}$ is used for supervision, with joints local rotation mean-squared error (MSE) and global position MSE, defined as:
\begin{equation}
    \mathcal{L}_{rot} = \text{rot-mse}(z^{1:N}, \hat{z}^{1:N}) \, , \mathcal{L}_{pos} = \text{pos-mse}(z^{1:N}, \hat{z}^{1:N}) \, .
\end{equation}

\begin{table*}
\centering
\setlength{\tabcolsep}{2.3mm}
\newcommand{\sz}{\hspace{2mm}}
\begin{center}
\begin{tabular}{lccccccc}
\toprule
Model & \makecell{MPJPE (cm) \\ up $|$ low }   & \makecell{MPJRE (radians) \\ up $|$ low}  & \makecell{MPJVE ($m^2$/$s^3$)\\ up $|$ low}   & \makecell{jitter (pred/gt) \\ up $|$ low} & \makecell{PC-loss \\  } \\
\midrule
\multicolumn{5}{c}{All Motions} \\
\midrule
AGRoL \cite{du2023agrol} (retrained)  &  1.87 $|$ 11.21   & 1.79 $|$ 2.30  &   10.27 $|$ 34.36  &  \textbf{2.55} $|$ \textbf{6.19} &  0.61 \\
\added{+} MPE &  1.78 $|$ 9.62  &  1.75 $|$ 2.23 &    9.94 $|$ 31.89 &   3.22 $|$12.66 & 0.39 \\
\added{+} MPE + SPC-decoder & 1.77 $|$ 9.30  &  1.75 $|$ \textbf{2.21} &   9.95 $|$ 32.41 & 3.11 $|$ 12.98 &  0.34 \\
\added{+} MPE + SPC-decoder + PC-loss &  1.78 $|$ 10.25  &  1.76 $|$ 2.25 &    9.96 $|$ 32.97 &    2.84 $|$ 9.43  &  0.48 \\
\added{+} MPE + SPC-decoder + SPC-loss &   \textbf{1.76} $|$ \textbf{9.27}  &  \textbf{1.74} $|$ \textbf{2.21} &    \textbf{9.78} $|$ \textbf{31.40} &  2.88 $|$ 9.47 &  \textbf{0.32}\\

\midrule
\multicolumn{5}{c}{Kicking} \\
\midrule
AGRoL \cite{du2023agrol} (retrained)  &  1.58 $|$ 10.21   &  1.63 $|$ 2.05  &   10.53 $|$ 41.10  &   \textbf{1.87} $|$ \textbf{3.32}  & 0.80 \\
\added{+} MPE &  1.49 $|$ 8.27  &  1.61 $|$ 2.01 &    10.22 $|$ 36.56 &    2.31 $|$ 7.6  & 0.39\\
\added{+} MPE + SPC-decoder &  \textbf{1.48} $|$ 7.97  &  1.61 $|$ 1.99 &   10.24 $|$ 37.76  &   2.23 $|$ 7.88  & 0.33\\
\added{+} MPE + SPC-decoder + PC-loss   &  1.50 $|$ 8.91  &  1.62 $|$ 2.02 &    10.26 $|$ 38.98 &   2.08 $|$ 5.70 & 0.51\\
\added{+} MPE + SPC-decoder + SPC-loss &   \textbf{1.48} $|$ \textbf{7.78}  &  \textbf{1.59} $|$\textbf{1.98}  &   \textbf{10.11} $|$ \textbf{36.42}  &     2.18 $|$ 5.93 & \textbf{0.28} \\
\bottomrule 
\centering
\end{tabular}
\end{center}
\vspace{-2.6em}
\caption{Ablation study on Mocap data test set with synthetically generated point clouds. 
\added{We compare the state-of-the-art AGRoL method to our MPE (Sec.~\ref{sec:mpe}) network, SPC decoder (Sec.~\ref{sec:spc}), and various self-supervised losses (Sec.~\ref{sec:loss}).}
Metrics are reported separately for the upper (up) and lower (low) body. Additional results for diverse motion protocols can be found in the supplementary material.}
\label{tbl:synthetic}
\end{table*}

\subsection{Self-supervised loss} 
\label{sec:loss}
Due to the nature of our input data, we could directly compute a self-supervised loss using the provided point cloud.
The most straightforward Point Cloud loss (PC-loss) is defined as an average distance between the point cloud and the closest face of the predicted body pose mesh, \ie we have to assume a default body shape \textit{in this simplest case}. 
However, mapping a pose to an unlabeled point cloud is too ambiguous and there is no guarantee that the right body parts would move toward the correct points. The predicted pose could simply deform to collide with the center of the point cloud while minimizing the PC-loss. 

Instead, we propose to leverage the SPC decoder to predict a semantic registration of the point cloud to formulate a novel Semantic Point Cloud Loss (SPC-loss). 
We define support for each joint $j$ as the sum of predicted probabilities from all points in the point cloud:
\begin{equation}
    \text{support}_j(l^t) = \sum_{i=1}^{P} l^t(i, j)  \, .
\end{equation}
For each body joint with support larger than $0.05 P$, which form a set $J^{*}$, we compute a spatial center that represents a noisy estimate of the joint position:
\begin{equation}
    \text{cen}_j(p^t, l^t) = \frac{1}{\text{support}_j(l^t)} \sum_{i=1}^{P} p_i^{t} \cdot l^t(i, j)   \, .
\end{equation}
This joint estimate is too noisy to be directly used by a run-time optimization methods (\eg inverse kinematics) as it lies on the surface of the body in most cases, and is largely affected by the sampling of the point cloud.
Thus, we introduce a threshold $\theta = $ 10 cm and penalize only distances over this threshold. 
The predicted joints are extracted by $\text{pos}_{j}(z^{t})$, and the final SPC-loss is \added{$\mathcal{L}_{spc} = $
\begin{equation}
\begin{aligned}
\frac{1}{N | J^{*}|} \sum_{t=1}^{N} \sum_{j \in J^{*}}  \max(0, \| \text{pos}_{j}(z^{t}) -  \text{cen}_j(p^t, l^t)\|_2 - \theta) \, .
\end{aligned}
\label{eq:spc}
\end{equation}}
As an example in Fig.~\ref{fig:header}, the noisy estimate of the joint position is represented as a cross, while the predicted joint position by the model is illustrated by a diamond. 
To preserve the realism of the initially synthesized pose, the body joints that are not visible to the sensors also need to be moved to a position that is consistent with the visible joints, for instance, to preserve a realistic balance. 

\boldparagraph{Fine-tuning on real data.}
Training only on synthetic point clouds does not capture the full, characteristic noise of the depth sensor and cannot fully generalize to real depth data, as demonstrated in Section~\ref{sec:experiments}, and an additional form of supervision is beneficial. 
Even though it is unlabeled, the real point cloud data can be used to supervise the training of the MPE network and learn an adaptation to the specific noise of the sensor.
During self-supervision on real data, we use the same self-supervised pose loss $\mathcal{L}_{spc}$~\eqref{eq:spc} on unlabeled depth data and minimize the distance of predicted centroids to the pose estimated by our MPE model.

\subsection{Training}
\label{sec:training}
The 3-Point synthesis model is pre-trained separately using the method described in \cite{du2023agrol}.
Then, it is frozen, and the SPC and MPE networks are jointly trained by minimizing a weighted sum of all previously defined loss terms:
\begin{equation}
    \mathcal{L} = w_{rot} \mathcal{L}_{rot} + w_{pos} \mathcal{L}_{pos} + w_{ce} \mathcal{L}_{ce}  + w_{spc}  \mathcal{L}_{spc} \, ,
\label{eq:loss}
\end{equation}
where weights are set empirically to $w_{rot} = 1$ (for error in radians), $w_{pos} = 0.01$ (for position error in meters), $w_{ce} = 0.1$, and $w_{spc} = 0.01$ (for error in meters).

The SPC and the MPE networks are jointly trained on a combination of synthetic Mocap and real data using ADAM optimizer~\cite{adam} for 20k iterations. 
The batch size for Mocap data is set to 128, and for real data, it is set to 32.  
The SPC-loss $\mathcal{L}_{spc}$ is averaged for both real and Mocap data. 
All other losses are computed only for the Mocap batch.
The learning rate is set to $0.0003$.
As the synthesis network is a diffusion model, we randomly sample new predictions for each batch during training of the PSC and MPE models to learn better robustness to the noise and diversity of the pose synthesis stage. 
We implement the MPE network as a simple MLP with 2M parameters, and the models are trained for 1 day on a cluster with 8 GPUs \added{NVIDIA} P100.

\begin{table}
\centering
\setlength{\tabcolsep}{0.6em}
\begin{center}
\begin{tabular}{lcccc}
\toprule
  & \added{AGRoL} & \added{SPC} & \added{MPE} & \added{Total} \\
\midrule
\added{Compute [GFLOPs]} & 122 & 89 & 2.5 & \added{213.5} \\
\added{Runtime [ms]} & 35 & 2.48 & 3.93 & \added{41.41} \\
\bottomrule 
\centering
\end{tabular}
\end{center}
\vspace{-3em}
\caption{\added{Compute and runtime of each step in our pipeline.}}
\label{tbl:compute}
\end{table}

\boldparagraph{Compute.}
AGRoL backbone compute is 122 GFLOPs for diffusing 5 DDIM steps. 
In addition, the SPC network compute is 89 GFLOPs, using the original PointNet implementation, but could be replaced by an efficient architecture for on-device integration such as PointNeXt~\cite{qian2022pointnext} reporting 7.2 GFLOPs for 3D semantic segmentation. 
The MPE network compute is 2.5 GFLOPs.
The inference time on a single P100 GPU is 2.48 ms for the SPC network and 3.93 ms for the MPE network. 
\added{Runtime of the full pipeline is 41.41 ms (Table~\ref{tbl:compute}), which corresponds to 25 fps, making our approach compatible with real-time usage on the device. }

\section{Experiments}
\label{sec:experiments}

\begin{figure*}
\centering
\includegraphics[width=1.0\linewidth]{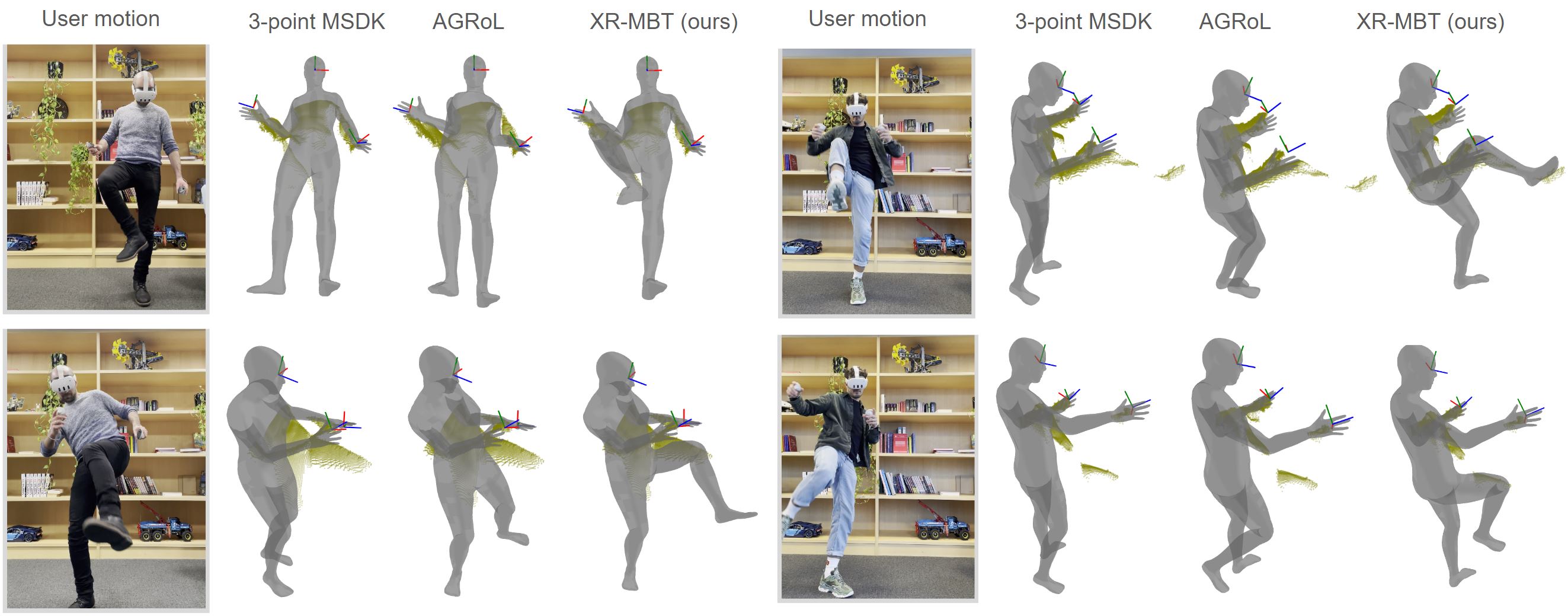}
\vspace{-2em}
\caption{\label{fig:qualitative_demo} Comparison of our XR-MBT method to other XR body tracking methods on real data. By leveraging multi-modal inputs in addition to the 3-Point data, our method is able to cover a large variety of lower body motions while preserving the plausibility of the pose for body parts outside the field of view of the depth sensor.}
\end{figure*}

\subsection{Datasets}

\boldparagraph{Mocap dataset.}
To ensure sufficient representation of lower body motions and to break down metrics by motion type, we replace the conventionally used AMASS dataset~\cite{amass2019} with our own Mocap dataset manually labeled with action types. 
Active lower body motions, where the tracked lower body is beneficial, are under-represented in the AMASS dataset~\cite{amass2019} and do not suit the task we propose to address in this paper.
To illustrate this, we extract BABEL~\cite{babel} action labels for AMASS and compute that it has only 2.7 hours of lower body active motions (out of 43 hours, 6\%).
In contrast, our Mocap dataset has 16 hours of such lower body active motions (out of 65 hours, 25\%).
The dataset entails 2964 motion capture sequences collected using a Vicon system from 330 subjects, 50\% male and 50\% female, with a high diversity of height and weight. 
Each sequence lasts on average 60 seconds and was designed to include a single action label such as "walking" or "knee strikes". 
The marker data was then fitted using a skeleton of 89 body joints. 
We retrain the AGRoL synthesis model on our dataset and skeleton model to account for these changes and report the metrics in Table~\ref{tbl:synthetic}.

\boldparagraph{Egocentric depth dataset.} 
We construct a real dataset of egocentric depth data taken from a real XR device. 
We collect a total of 203 sequences from 11 subjects executing diverse body motion protocols while using a Quest 3 VR headset. 
For each sequence, we record data from the 3-Point tracking signal of the controllers and headset tracking, as well as the point clouds from the native depth sensing system of the device. 
We additionally collect 20 test sequences that are not used during training, and where user motions are additionally captured from a mobile phone camera for qualitative evaluation. 
As the collected point cloud does not only cover the body but the entire scene, we filter the body point cloud using simple proximity and density assumptions to retrieve only the body parts. 
Examples of the obtained point clouds are visible in Figs.~\ref{fig:depth_data_demo} and \ref{fig:qualitative_demo}.

\subsection{Experiments on synthetic data}

To conduct an ablation study in a controlled environment, we first evaluate each component of our method on Mocap data with simulated point clouds. 
Such evaluation is commonly encountered in the XR body tracking literature as there is no available dataset of real 3-Point inputs with body GT. 
We individually train and evaluate different components of our method (Table~\ref{tbl:synthetic}). 
We reuse the metrics definition from~\cite{du2023agrol}, with the exception of the jitter metric that we report as the ratio between the predicted and ground-truth jitter for better readability. 
Additionally, we report the PC-loss (the point cloud loss without registration).

\begin{figure}[t]
\centering
\includegraphics[width=0.99\linewidth]{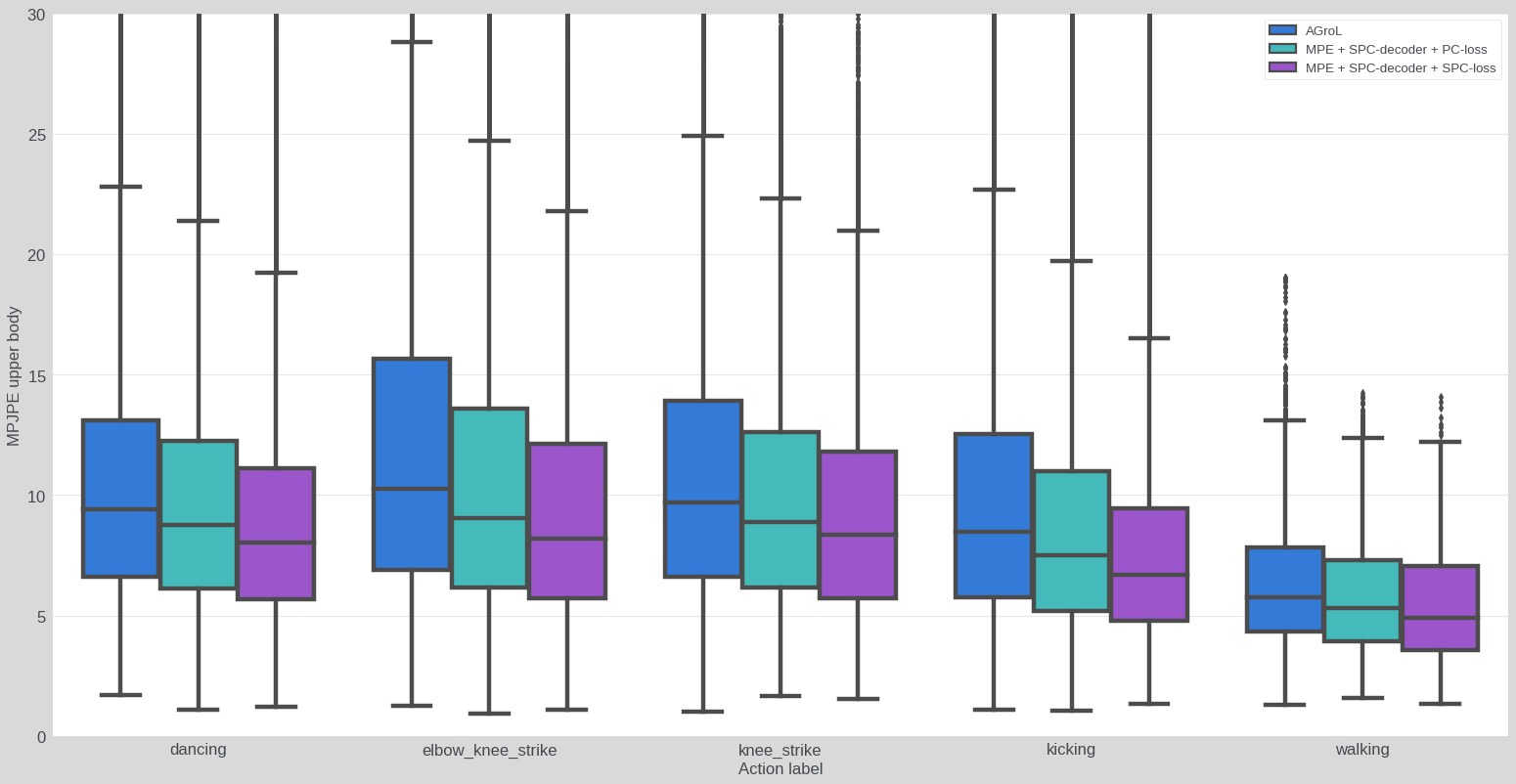}
\vspace{-0.8em}
\caption{\label{fig:motion_label_boxplots} MPJPE for lower body computed on the Mocap test set with different action labels. We compare AGRoL~\cite{du2023agrol} to our method trained with PC-loss or SPC-loss~\eqref{eq:spc}.}
\end{figure}

Several observations can be made from these results. 
First, using an SPC decoder module improves the lower body accuracy compared to training the MPE network using only a PointNet encoder. 
This demonstrates the importance of learning a semantically meaningful point cloud registration as input to the MPE network. 
Second, using the simple PC-loss for self-supervised training of the MPE network leads to an increased error compared to using only pose supervision. 
Indeed, minimizing the distance between the predicted pose and an unregistered point cloud is too ambiguous as the closest body pose vertex might belong to a different body part than the one visible to the depth sensor. 
However, the introduction of the SPC-loss $\mathcal{L}_{spc}$~\eqref{eq:spc} leveraging the SPC decoder outputs allows to improve the quality of self-supervision, decreasing the lower body pose error for kicking motions from 8.91 cm to 7.79 cm. 
Note that our motion labels are attributed at the sequence level and not frames. 
As a result, the kicking frames are still underrepresented in the kicking sequences (10\%). 
Thus, these metrics underestimate the margin in quality between our lower body estimate and conventional 3-Point methods, as shown by the qualitative evaluation in the following section.
We also report the distribution of the lower body pose error for different motions in Fig.~\ref{fig:motion_label_boxplots} and show that the self-supervised SPC-loss generalizes well to diverse types of action labels. 
Even though it only brings a marginal improvement in such synthetic experiments, formulating self-supervision to learn adaptation to the depth sensor characteristics is key when training and evaluating with real data, as we demonstrate in the following section. 
Finally, computing the PC-loss on the synthetic point clouds allows to verify that the SPC-loss learns to leverage the point cloud data better. 

The improvement in walking metrics can intuitively be explained by the following factors:
\textbf{(1)} Due to the large field of view of the depth sensor, the lower body and, in particular, the upper leg and knee are frequently visible in the point cloud (even when not looking down), providing strong information to improve leg position accuracy. 
\textbf{(2)} When a leg is not visible to the depth sensors, this also constitutes a relevant signal to the model that this leg should not be stepping forward, which can improve the stepping cycle accuracy. 

Compared to the 3-Point synthesis task, our method leads to a higher jitter level. 
While the jitter increase is controlled for the upper body, going from 2.55 to 2.88 on average, the increase in the lower body jitter is substantially more significant, going from 6.19 to 9.47. 
To some extent, this is expected as the lower body is no longer synthesized but tracked. 
The leg joints are activated more often, particularly for the motion types that we use. 
An excessive jitter could be corrected by using a temporal model or IK constraints at run-time to regularize body joint activations. 

\boldparagraph{Extended ablation study results.}
Table~\ref{tbl:synthetic_extended} shows an extended version of the ablation study. 
The results on motions with body parts well covered by the depth signal show the benefits of training the MPE network to grasp and leverage semantic information. On the one hand, this is achieved by training an SPC decoder to predict semantics and by adding our novel SPC-loss. 
However, note that only part of the sequences contains active motions and that our vanilla MPE solution is sufficient for more common and symmetric motion categories, such as bare walking already. The previous state-of-the-art AGRoL is outperformed in all cases.

\subsection{Experiments on real data}

Finally, we evaluate our approach against several state-of-the-art solutions in the field of XR body tracking. 
In particular, we use our unlabeled 3-Point egocentric depth dataset to compare the proposed XR-MBT architecture against AGRoL~\cite{du2023agrol} and the commercially available 3-Point MSDK~\cite{meta_msdk}. 
We could not include the MSDK model in the synthetic experiments, as it is only available as a Unity package, making it impossible to control the training set. 

\begin{figure*}
\centering
\includegraphics[width=0.99\linewidth]{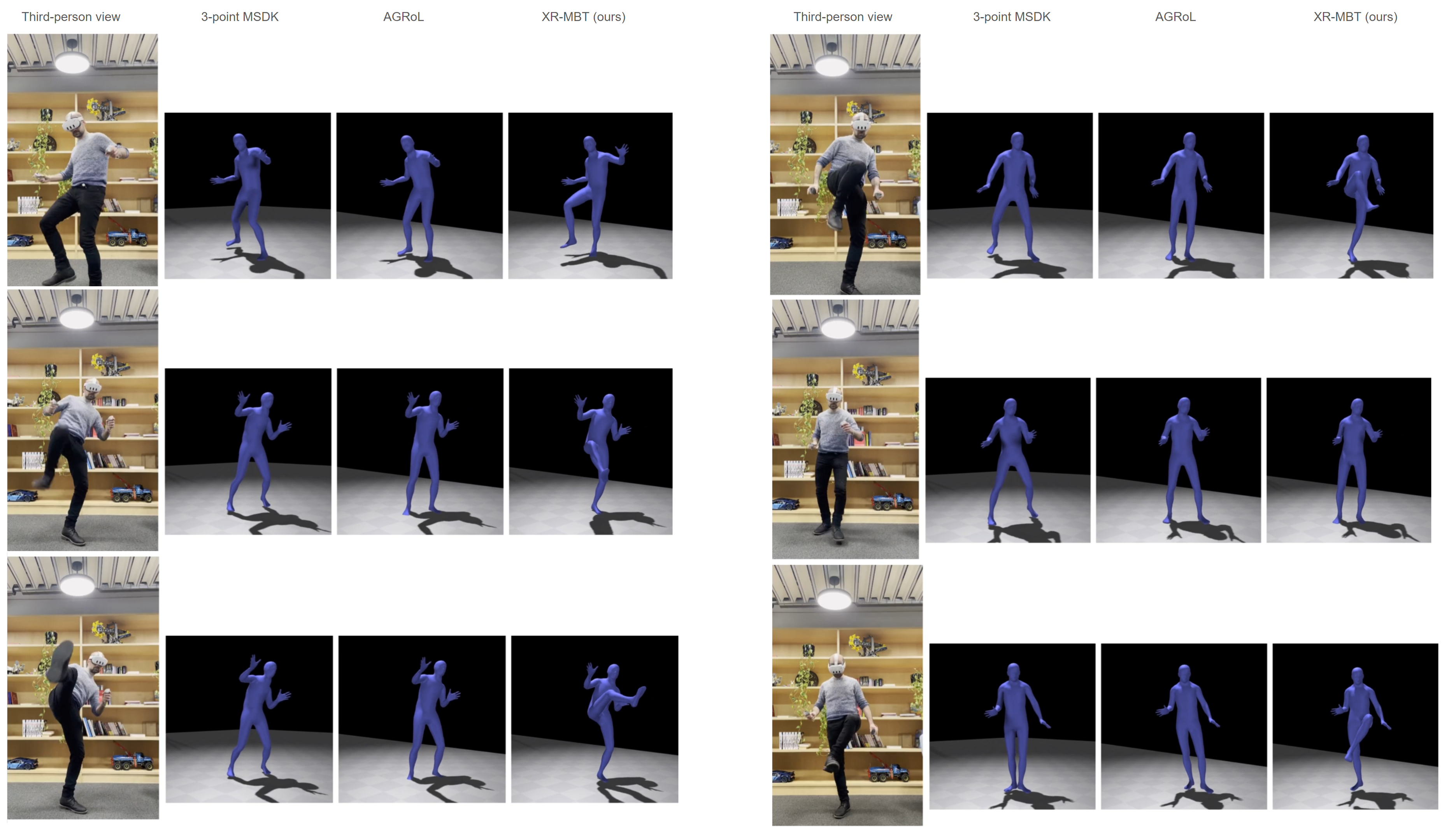}
\caption{\added{Additional results on real data.}}
\label{fig:new_real_data}
\end{figure*}

\begin{figure*}
\centering
\includegraphics[width=0.99\linewidth]{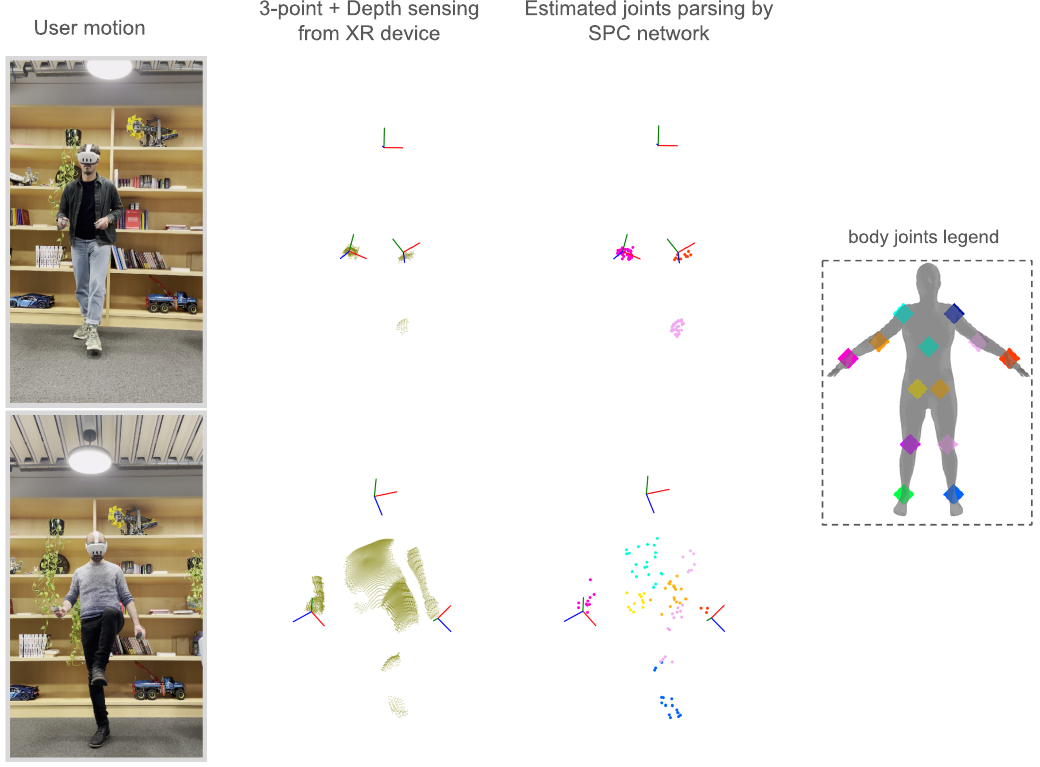}
\caption{Additional point cloud labeling results.}
\label{fig:extended_visual_pc_results}
\end{figure*}

For qualitative evaluation of the SPC network, we use real point cloud data from our test sequences and visualize the body classification of the SPC decoder. 
These results are visible in Fig.~\ref{fig:depth_data_demo}, Fig.~\ref{fig:extended_visual_pc_results}, and Fig.~\ref{fig:new_real_data}. 
Even when the body point cloud obtained from real depth data is partial due to the self-occlusion of the body and the limited field of view of the depth sensor, the proposed SPC decoder learns to classify various body joints accurately. 
This indicates that the SPC network learns a meaningful registration of the body point cloud to supervise the MPE network training.

In addition, we run different XR body tracking models on \added{real test data sequences} and visually compare the results with reference videos.
The results are visible in Fig.~\ref{fig:qualitative_demo}.
It can be observed that the refinement step successfully learns to transition from a synthesized pose to a tracked pose that matches the point cloud signal. 
This transition is more complex than moving detected body joints toward the point cloud, and we can see in Figs.~\ref{fig:header} and \ref{fig:qualitative_demo} that the MPE network successfully learns to move the leg not visible to the depth sensor to maintain the character balance and the overall plausibility of the motion. 
It can also be observed that using the point cloud information allows a better pose of the upper body joints visible to the sensor, such as the torso and elbows. 
By leveraging multi-modal inputs in addition to the 3-Point data, our method can cover a largely increased range of motions, such as kicking, knee strike and football, while preserving the realism of the motions when body parts are outside the field of view.

\begin{table}[t]
\centering
\setlength{\tabcolsep}{1.3mm}
\newcommand{\sz}{\hspace{1mm}}
\begin{center}
\begin{tabular}{l@{\hskip 0.1em}ccccc}
\toprule
Motion &  MSDK~\cite{meta_msdk} & AGRoL~\cite{du2023agrol} & XR-MBT & + real \\
\midrule
\added{walking}  &  \added{7.16}  & \added{7.44} &  \added{4.67} &  \added{4.39} \\
kicking &  7.60  & 6.41 &  5.01 & 3.74 \\
football &  1.70  & 1.90 &  1.79 & 1.22   \\
beatsaber &  6.77  & 5.39 &  4.70 & 2.90    \\
knee strikes &  1.16  & 1.13 & 1.04 &  0.93    \\
lift legs &  2.45  & 2.19 &  1.67 & 1.64    \\
\midrule
average &  4.12  & 3.74 & 3.02 &  2.58   \\

\bottomrule 
\centering
\end{tabular}
\end{center}
\vspace{-2.9em}
\caption{Point cloud error (PC-loss, in cm) computed on real test data for different actions. These results demonstrate that training with self-supervision on real data allows to learn adaptation to the real depth sensors, leading to a reduced error compared to training on synthetic point cloud data.}
\label{tbl:point_cloud_metrics}
\end{table}

\boldparagraph{Limitations.}
One apparent limitation of our approach is the absence of a representation of the user's body shape. 
Indeed, even though we cannot assume that the user's body shape has been previously registered, an explicit representation of the user's shape could be beneficial to use as input to our approach. To that extent, solving the body shape of the user at run-time given the partial point cloud data could constitute a natural extension of our method and help disambiguate the point cloud and refine the pose further. 
Another limitation is that we assume that the user does not interact with any large objects, which could create interference in the point cloud registration. 
\added{Moreover, motion quality could be improved by using human motion priors~\cite{humor}.}

\begin{figure*}
\centering
\begin{tabular}{ccc}
\includegraphics[height=0.35\linewidth]{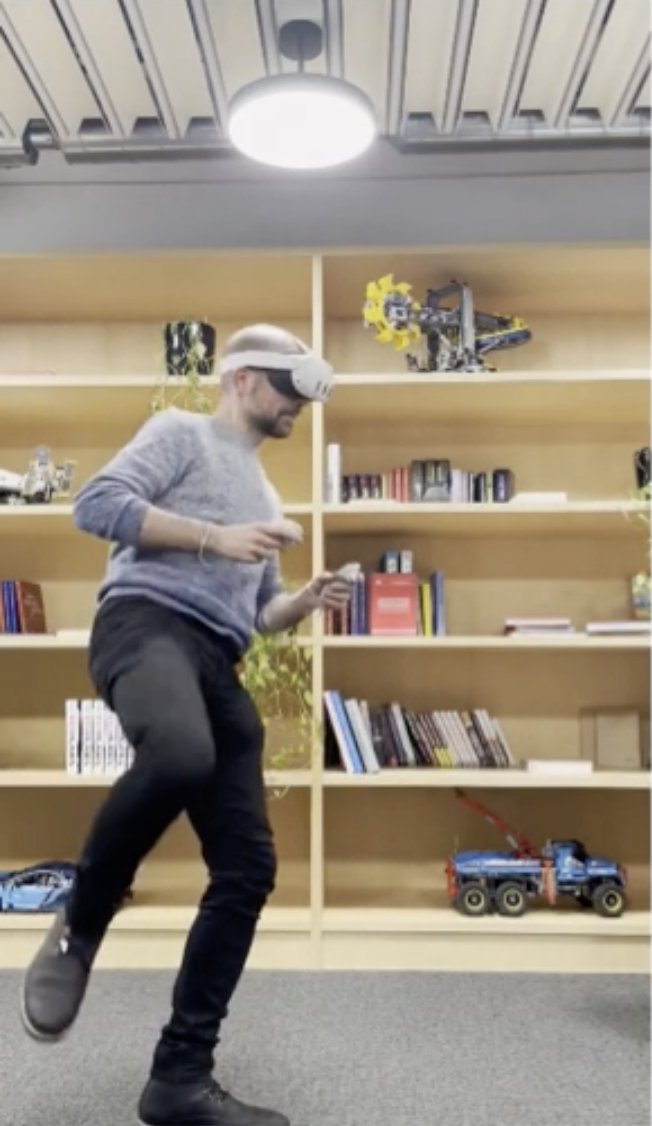} & 
\includegraphics[height=0.35\linewidth]{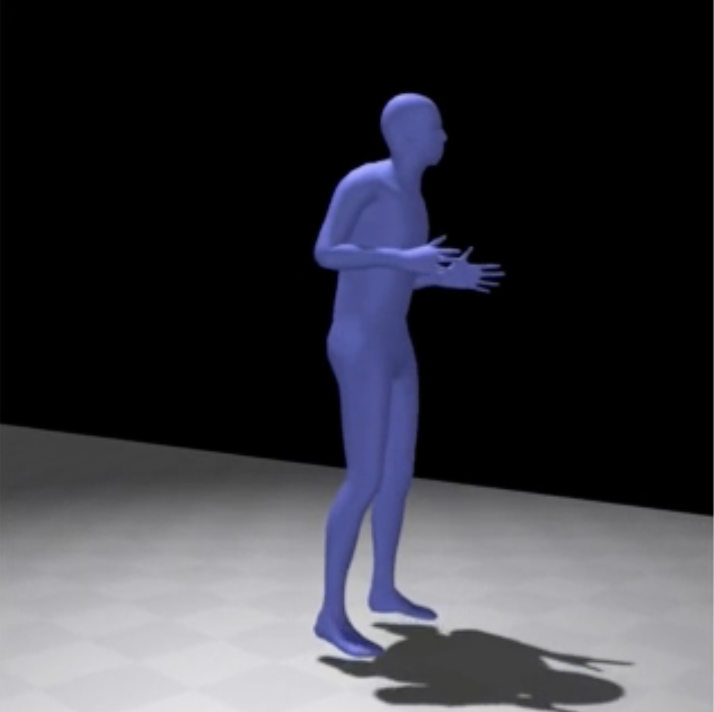} &
\includegraphics[height=0.35\linewidth]{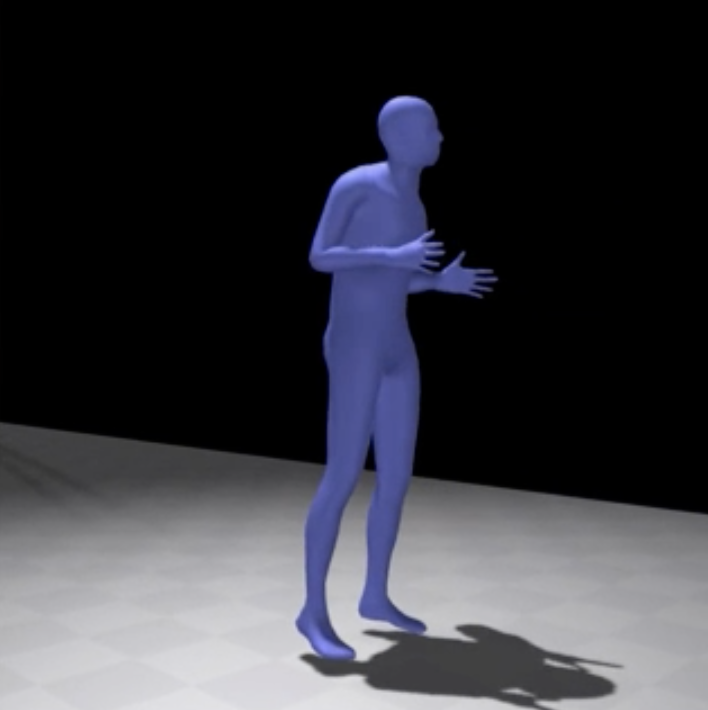} \\
Third-person view & AGRoL & XR-MBT (ours) \\
\end{tabular}
\vspace{-1em}
\caption{\added{A failure in case of limited field of view, \ie the lifted right leg is not visible to the XR device.}}
\label{fig:fail}
\end{figure*}

\boldparagraph{Failure cases.}
\added{Failure cases occur when lower body motion happens outside the field of view of the sensor, such as in Fig.~\ref{fig:fail}.}

\begin{table*}
\centering
\setlength{\tabcolsep}{2.3mm}
\newcommand{\sz}{\hspace{5mm}}
\begin{center}
\begin{tabular}{lccccccc}
\toprule
\footnotesize
Model & \makecell{MPJPE (cm) \\ up $|$ low }   & \makecell{MPJRE (radians) \\ up $|$ low}  & \makecell{MPJVE ($m^2$/$s^3$)\\ up $|$ low}   & \makecell{jitter (pred/gt) \\ up $|$ low} & \makecell{PC-loss \\  } \\
\midrule
\multicolumn{5}{c}{All Motions} \\
\midrule
AGRoL (retrained)  &  1.87 $|$ 11.21   & 1.79 $|$ 2.30  &   10.27 $|$ 34.36  &  \textbf{2.55} $|$ \textbf{6.19} &  0.61 \\
MPE &  1.78 $|$ 9.62  &  1.75 $|$ 2.23 &    9.94 $|$ 31.89 &   3.22 $|$12.66 & 0.39 \\
MPE + SPC-decoder & 1.77 $|$ 9.30  &  1.75 $|$ \textbf{2.21} &   9.95 $|$ 32.41 & 3.11 $|$ 12.98 &  0.34 \\
MPE + SPC-decoder + PC-loss &  1.78 $|$ 10.25  &  1.76 $|$ 2.25 &    9.96 $|$ 32.97 &    2.84 $|$ 9.43  &  0.48 \\
MPE + SPC-decoder + SPC-loss &   \textbf{1.76} $|$ \textbf{9.27}  &  \textbf{1.74} $|$ \textbf{2.21} &    \textbf{9.78} $|$ \textbf{31.40} &  2.88 $|$ 9.47 &  \textbf{0.32}\\

\midrule
\multicolumn{5}{c}{Kicking} \\
\midrule
AGRoL (retrained)  &  1.58 $|$ 10.21   &  1.63 $|$ 2.05  &   10.28 $|$ 39.22  &   \textbf{1.87} $|$ \textbf{3.32}  & 0.80 \\
MPE &  1.49 $|$ 8.27  &  1.61 $|$ 2.01 &    10.11 $|$ 36.56 &    2.31 $|$ 7.6  & 0.39\\
MPE + SPC-decoder &  \textbf{1.48} $|$ 7.97  &  1.61 $|$ 1.99 &   10.25 $|$ 37.77  &   2.23 $|$ 7.88  & 0.33\\
MPE + SPC-decoder + PC-loss   &  1.50 $|$ 8.91  &  1.62 $|$ 2.02 &    10.26 $|$ 38.98 &   2.08 $|$ 5.70 & 0.51\\
MPE + SPC-decoder + SPC-loss &   \textbf{1.48} $|$ \textbf{7.78}  &  \textbf{1.59} $|$\textbf{1.98}  &   \textbf{10.11} $|$ \textbf{36.42}  &     2.18 $|$ 5.93 & \textbf{0.28} \\

\midrule
\multicolumn{5}{c}{Knee strikes} \\
\midrule
AGRoL (retrained)  &  1.63 $|$ 10.67   &  1.58 $|$ 2.03  &   9.97 $|$39.97  &   \textbf{1.91} $|$ \textbf{3.34}  & 0.32 \\
MPE &  1.57 $|$ \textbf{9.36}  &  1.55 $|$ 1.95 &    \textbf{9.83} $|$ \textbf{38.56} &    2.28 $|$ 7.81  & \textbf{0.25} \\
MPE + SPC-decoder &  \textbf{1.51} $|$ 9.38  &  \textbf{1.54} $|$ \textbf{1.93} &   9.89 $|$ 38.79  &   2.28 $|$ 7.81   & 0.30 \\
MPE + SPC-decoder + PC-loss   &  1.53 $|$ 10.16  &  1.56 $|$ 2.00 &   9.91 $|$ 40.68 &   2.21 $|$ 5.96 & 0.33 \\
MPE + SPC-decoder + SPC-loss &   1.57 $|$ \textbf{9.36}  &  1.57 $|$1.95  &   \textbf{9.83} $|$ \textbf{38.56}  & 2.30 $|$ 7.18 & \textbf{0.25} \\

\midrule
\multicolumn{5}{c}{Elbow knee strikes} \\
\midrule
AGRoL (retrained)  &  2.14 $|$ 11.82   &  1.87 $|$ 2.29  &   2.09 $|$ 5.54  &   \textbf{1.56} $|$ \textbf{2.50}  & 0.35 \\
MPE &  2.09 $|$ 10.27  &  1.85 $|$ \textbf{2.23} &    \textbf{2.08} $|$ \textbf{5.35} &    1.75 $|$ 5.27  & \textbf{0.24} \\
MPE + SPC-decoder &  \textbf{2.06} $|$ 10.38  &  1.82 $|$ 2.37 &   2.10 $|$ 5.54  &   1.80 $|$ 5.86  & 0.26 \\
MPE + SPC-decoder + PC-loss   &  2.08 $|$ 11.36  &  1.83 $|$ 2.41 &   2.10 $|$ 5.64 &   1.66 $|$ 4.32 & 0.35 \\
MPE + SPC-decoder + SPC-loss &   2.09 $|$ \textbf{10.17}  &  \textbf{1.83} $|$ 2.34  &   \textbf{2.08} $|$ \textbf{5.35}  & 1.75 $|$ 5.27 & 0.26 \\

\midrule
\multicolumn{5}{c}{Walking} \\
\midrule
AGRoL (retrained)  &  1.33 $|$ 5.77   &  1.45 $|$ 1.63  &   7.41 $|$ 23.65  &   \textbf{1.79} $|$ \textbf{2.41}  & 0.82 \\
MPE &  1.27 $|$ \textbf{5.42}  &  1.44 $|$ \textbf{1.59} &    \textbf{7.21} $|$ \textbf{23.19} &    2.09 $|$ 4.80  & \textbf{0.62} \\
MPE + SPC-decoder &  \textbf{1.22} $|$ 5.61  &  \textbf{1.42} $|$ 1.64 &   7.36 $|$ 24.00  &   2.19 $|$ 6.10  & 0.65 \\
MPE + SPC-decoder + PC-loss   &  1.23 $|$ 5.82  &  1.43 $|$ 1.66 &   7.35 $|$ 24.25 &   2.00 $|$ 4.07 & 0.74 \\
MPE + SPC-decoder + SPC-loss &  1.25 $|$ 5.44  &  1.45 $|$ 1.70  &   7.22 $|$ \textbf{23.19}  & 2.09 $|$ 4.80 & \textbf{0.62} \\

\bottomrule 
\centering
\end{tabular}
\end{center}
\vspace{-2em}
\caption{Ablation study on the Mocap data test set with synthetically generated point clouds. Metrics are reported separately for the upper (up) and lower (low) body.}
\label{tbl:synthetic_extended}
\end{table*}

\section{Conclusion}

In this work, we introduced for the first time the use of depth sensors in the XR body tracking task.
While these sensors provide a useful signal to drive the body, the unregistered point cloud is ambiguous and cannot be directly used to drive body tracking.
We proposed a solution by learning to predict a semantic registration of the egocentric point cloud.
We showed that this learned point cloud registration could be leveraged to formulate a self-supervised loss, which in turn enabled training a multi-modal body tracking model on real unlabeled depth data.
As a result, our architecture is capable of automatically transitioning between synthesis and tracking depending on the body parts visible to the sensors in order to track the lower body and always guarantee a plausible pose and an authentic body presence in XR.
We believe that this work opens novel perspectives in the field of XR body tracking by exploiting readily available depth data in an efficient and effective manner.

{\small
\bibliographystyle{ieee_fullname}
\bibliography{main}
}

\end{document}